\documentclass{article}

    \usepackage[preprint]{neurips_2025}

\usepackage{multirow}
\usepackage[utf8]{inputenc} 
\usepackage[T1]{fontenc}    
\usepackage{hyperref}       
\usepackage{url}            
\usepackage{booktabs}       
\usepackage{amsfonts}       
\usepackage{nicefrac}       
\usepackage{microtype}      
\usepackage{xcolor}         
\usepackage{graphicx}
\usepackage{amsmath} 
\usepackage{caption}
\usepackage{xspace}

\usepackage{float}

\makeatletter
\DeclareRobustCommand\onedot{\futurelet\@let@token\@onedot}
\def\@onedot{\ifx\@let@token.\else.\null\fi\xspace}

\def\eg{\emph{e.g}\onedot} 
\def\ie{\emph{i.e}\onedot}

\newcommand{\blue}[1]{\underline{#1}}

\title{Implicit Neural Representation for Video Restoration}

\author{%
  Mary Aiyetigbo\thanks{Clemson University, School of Computing} \\
  \And
  Wanqi Yuan\footnotemark[1] \\
  \And
  Feng Luo\footnotemark[1] \\
  \And
  Nianyi Li\footnotemark[1] \\
}

\begin{document}

\maketitle

\begin{abstract}
High-resolution (HR) videos play a crucial role in many computer vision applications.  Although existing video restoration (VR) methods can significantly enhance video quality by exploiting temporal information across video frames, they are typically trained for fixed upscaling factors and lack the flexibility to handle scales or degradations beyond their training distribution. In this paper, we introduce VR-INR, a novel video restoration approach based on Implicit Neural Representations (INRs) that is trained only on a single upscaling factor ($\times 4$) but generalizes effectively to arbitrary, unseen super-resolution scales at test time. Notably, VR-INR also performs zero-shot denoising on noisy input, despite never having seen noisy data during training. Our method employs a hierarchical spatial-temporal-texture encoding framework coupled with multi-resolution implicit hash encoding, enabling adaptive decoding of high-resolution and noise-suppressed frames from low-resolution inputs at any desired magnification. Experimental results show that VR‑INR consistently maintains high-quality reconstructions at unseen scales and noise during training, significantly outperforming state‑of‑the‑art approaches in sharpness, detail preservation, and denoising efficacy. The project page is available at \href{https://maryaiyetigbo.github.io/VRINR/}{https://maryaiyetigbo.github.io/VRINR/}
\end{abstract}   
\section{Introduction}
High-resolution (HR) videos are essential for numerous computer vision applications, including surveillance \cite{cristani2004distilling,zhang2014super}, medical imaging \cite{greenspan2009super,li2021review}, and multimedia entertainment \cite{kappeler2016video}. However, capturing high-resolution data is often constrained by hardware limitations, bandwidth, and storage considerations. 
Video restoration techniques, which encompass both super-resolution and denoising, aim to reconstruct high-quality frames from degraded low-resolutions (LR) sequences and thus have become a critical research direction \cite{park2003super}. 
%
\begin{figure}[t]
    \centering
    \includegraphics[width=.7\linewidth]{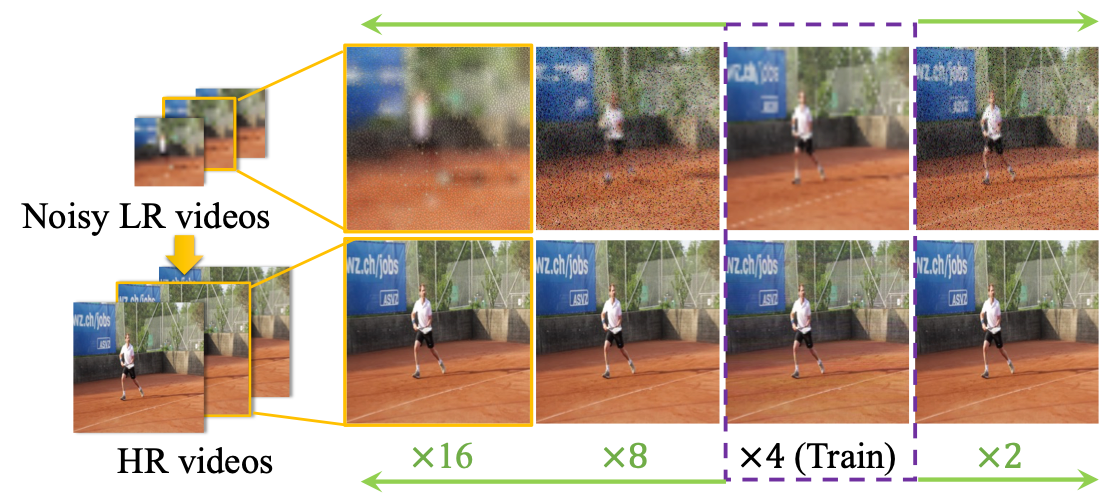}
    \caption{VR‑INR demonstrates robust scale generalization and zero‑shot denoising for video restoration. Although trained only on clean LR–HR pairs at \(\times4\), VR‑INR generalizes to arbitrary unseen scales (e.g., \(\times2\), \(\times8\), \(\times16\)) and removes noise from degraded inputs without any noise‑specific training, producing high‑quality restoration across scales and noise levels. Bottom row: VR‑INR outputs; green labels denote out‑of‑distribution scales.}

    \label{fig:teaser_fig}
    \vspace{-10pt}
\end{figure}
%

Modern video restoration techniques have significantly improved by utilizing temporal information across frames. Traditional methods often depend on explicit motion estimation, such as optical flow, to align frames before reconstruction ~\cite{liang2024vrt, liang2022recurrent, shi2022rethinking, chan2021basicvsr, xu2023video, tian2020tdan, wang2019edvr, jo2018deep, yang2025motion, lyu2024stadnet, zhou2024upscale, chen2024learning, zhang2024feature}. While effective, these approaches can be computationally intensive and may falter under complex motion or occlusion scenarios~\cite{wang2019edvr,ulyanov2018deep}.
To address these challenges, recent advancements have shifted towards implicit alignment strategies~\cite{wang2019edvr,liang2024vrt}. These methods employ advanced architectures, \eg, deformable convolutions and transformers, to capture temporal dependencies without direct motion estimation, enhancing both consistency and fidelity. Generative models, including GANs and diffusion models, have further elevated perceptual quality by synthesizing realistic textures~\cite{liu2025vsrdiff, jiang2024tempdiff, yuan2024inflation}. However, many of these networks are tailored to fixed upscaling factors (\eg $\times 4$) and require retraining to handle different scales or degradations like noise.
Implicit Neural Representations (INRs) present a flexible alternative for video restoration \cite{chen2021liif}. By modeling videos as continuous functions parameterized by neural networks, INRs inherently support arbitrary resolution queries \cite{chen2021liif}. Early applications in image super‑resolution demonstrated the potential of coordinate‑based networks for continuous upsampling \cite{chen2021liif}. In the video domain, VideoINR \cite{chen2022videoinr} and NeRV \cite{chen2021nerv} enabled arbitrary spatial scaling and frame interpolation within a single implicit framework. Moreover, recent studies have applied INRs to unsupervised video denoising via per‑video fitting \cite{aiyetigbo2024unsupervised}. Despite these advances, existing INR models still struggle to jointly generalize across both unseen scales and unseen degradations within a single trained network. 

In this paper, we propose VR‑INR, an implicit neural representation framework designed for video restoration that (i) is trained only on clean data at a single $\times4$ super‑resolution scale, (ii) generalizes to arbitrary unseen upscaling factors (\eg $\times2$,$\times8$,$\times16$) without retraining, and (iii) performs implicit denoising on noisy inputs at inference despite never having been trained on noisy videos. VR‑INR integrates a hierarchical spatial–temporal–texture encoder with a multi‑resolution hash embedding module to reconstruct high‑fidelity frames seamlessly, avoiding explicit motion estimation. We also introduce a pixel‑error amplified loss tailored to coordinate‑based restoration, which emphasizes high‑frequency residuals and reduces artifacts.
Our main contributions are as follows:
\begin{itemize} 
\item A novel unified video restoration framework that can address both arbitrary output resolution and denoising in a zero-shot manner, without explicit optical flow/motion estimation.
\item A novel hierarchical grid-based encoding strategy that leverages multi-resolution hash embeddings to construct an efficient implicit neural representation for video restoration.
\item A novel pixel-error amplified loss tailored for coordinate-based reconstruction and restoration framework to reduce reconstruction artifacts. 
\end{itemize}

\section{Related Work}

\paragraph{Learning-Based Video Restoration (VR).}
Traditional video restoration techniques often target specific degradation types, \eg noise, blur, or compression artifacts, using models tailored to each. However, real-world scenarios frequently involve multiple, time-varying unknown degradations, posing significant challenges to these specialized approaches.
Recent advances have introduced unified frameworks capable of addressing various degradations within a single model. For instance, AverNet \cite{zhao2024avernet} proposes an All-in-one Video Restoration Network to restore videos afflicted by multiple, unknown, and temporally varying degradations without prior knowledge of the degradation types. While effective, AverNet depends on explicit flow estimation and carefully crafted prompts, and it \textit{does not support arbitrary spatial scaling or zero‑shot denoising}. Recent advances in Sliding-window methods \cite{shi2015convolutional, chan2021basicvsr, chan2022basicvsr++, liang2022recurrent, haris2019recurrent, huang2017video, sajjadi2018frame, shi2022rethinking, yi2021omniscient}, including EDVR~\cite{wang2019edvr}, BasicVSR++~\cite{chan2021basicvsr}, VRT~\cite{liang2024vrt} implements implicit alignment strategies using deformable convolutions and transformer-based architectures, thus improving temporal consistency and reconstruction quality. RVRT \cite{liang2022recurrent} balances efficiency and effectiveness by integrating local parallel processing within a global recurrent framework, using guided deformable attention to align and aggregate features across different clips. IART \cite{xu2024enhancing} proposes an implicit resampling-based alignment, encoding sampling positions with sinusoidal positional encoding while utilizing a coordinate network and window-based cross-attention for feature reconstruction. 
SAVSR \cite{li2024savsr} proposes an iterative bi-directional architecture with scale-aware convolutions and a spatio-temporal adaptive upsampling module to achieve arbitrary-scale video super-resolution using a single model.
While these approaches signify a shift towards more adaptable and generalized video restoration models, many of them still require explicit training on each degradation type, limiting their adaptability to unforeseen degradation combinations.

\vspace{3pt}
\noindent \textbf{Implicit Neural Representations for VR.} 
Implicit Neural Representations (INRs) have recently gained prominence by modeling signals as continuous functions parameterized by neural networks, thereby inherently supporting arbitrary resolution generation. Early applications of INRs in super-resolution primarily targeted image-based tasks, as demonstrated by methods such as LIIF\cite{chen2021liif}, SIREN\cite{sitzmann2020siren}, and Fourier Features~\cite{tancik2020fourier}. These methods effectively capture intricate spatial details but often lack temporal modeling capabilities crucial for high-quality video reconstruction.
Recently, INR approaches have been extended to video super-resolution, with significant advancements including NeRV\cite{chen2021nerv}, HNeRV\cite{han2022hnerv}, and VideoINR~\cite{chen2022videoinr}. NeRV introduces a neural representation that directly encodes an entire video into a compact neural network, enabling efficient video reconstruction without explicit temporal modeling. VideoINR, in contrast, learns a continuous function that performs both spatial and temporal super-resolution, allowing for frame interpolation and reconstruction at arbitrary resolutions and time steps. Beyond super-resolution, INRs have also shown promise for denoising. Aiyetigbo \emph{et al.} \cite{aiyetigbo2024unsupervised} apply per‐video fitting of a coordinate‐based network to perform unsupervised video denoising.
%
However, existing INR-based VR methods still face challenges related to efficiently encoding fine-grained textures and maintaining reconstruction fidelity under dynamic and complex scenarios. To overcome these limitations, our method, VR-INR, proposes a novel hierarchical texture encoding framework combined with multi-resolution hash encoding~\cite{mueller2022instant}. Our approach significantly enhances reconstruction accuracy, temporal consistency, and computational efficiency for arbitrary-scale video super-resolution, effectively addressing the shortcomings of prior methods.

\begin{figure*}[t!]
    \centering
    \includegraphics[width=.9\textwidth]{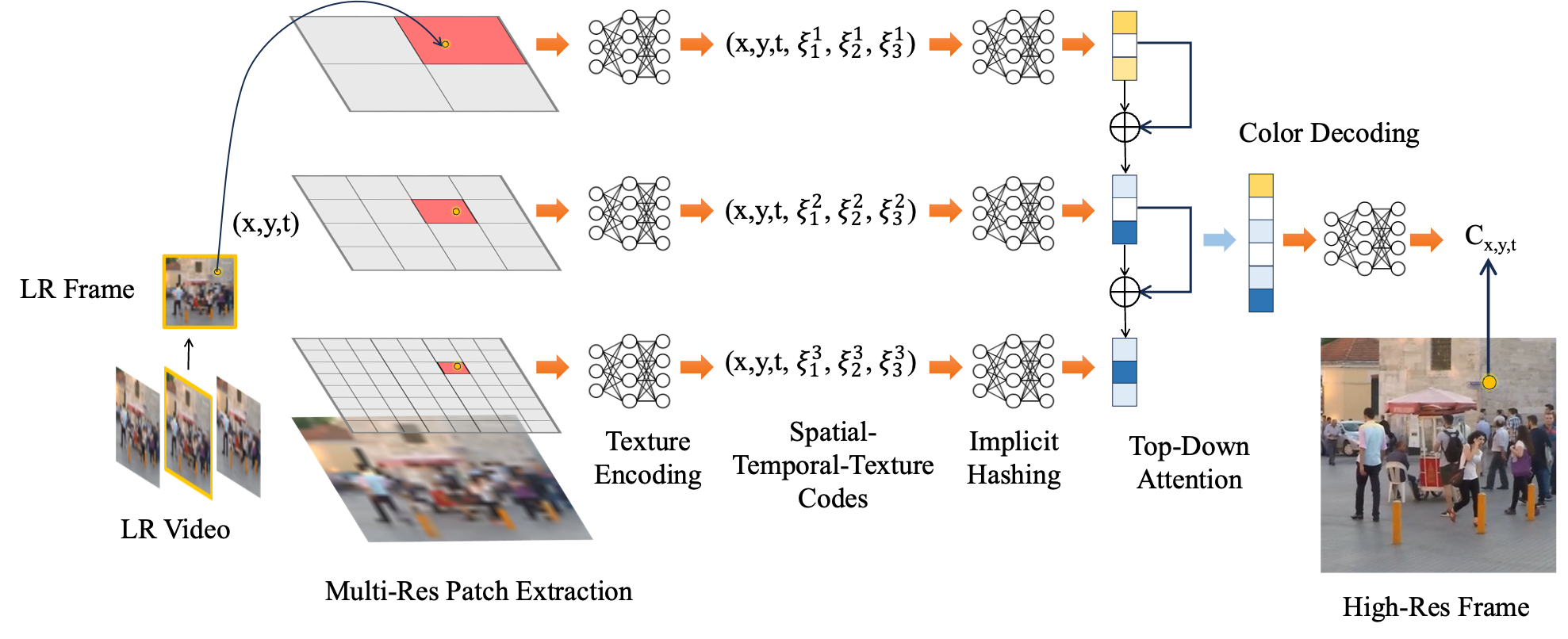}
    \caption{VR-INR training pipeline. Local patches are extracted at multiple resolutions and processed by MLPs to generate feature vectors. These vectors are concatenated, refined via a top-down attention mechanism, and fed into an MLP to predict the RGB value, resulting in the super-resolved output.}
    \label{fig:pipeline}
    \vspace{-15pt}
\end{figure*}

\section{Method}



We propose VR-INR, a novel video restoration approach based on Implicit Neural Representations. VR-INR is trained only on clean data for super-resolution but generalizes effectively to arbitrary, unseen super-resolution scales at test time. An overview of VR-INR training is shown in Fig.~\ref{fig:pipeline}. Given an input sequence of low-resolution (LR) video: $\{\mathbf{I}^{\text{LR}}_{t}|t = 1, 2, \ldots, T\}$ (where $T$ is the total number of frames, and $\mathbf{I}^{\text{LR}}_{t}$ represents a LR frame in the video) and a high-resolution grid $\mathbf{r}^{\text{HR}}\in\mathbb{R}^2$ specifying the spatial coordinates, VR-INR aims to produce high-resolution (HR) videos $\{\mathbf{I}^{\text{HR}}_{t}|t = 1, 2, \ldots, T\}$. First, we employ hierarchical texture encoding network (Section~\ref{sec:texture_encoding}) to extract and encode multi-scale local patches into spatial-temporal-texture feature representations $\mathbf{F}_{\text{STT}}$. For each target high-resolution coordinate $\mathbf{r}^{\text{HR}}$ at frame $t$, we retrieve a compact set of neighboring feature vectors from a spatial hash table using implicit hashing (Section~\ref{sec:implicit_hashing}), and efficiently interpolate these vectors using adaptively learned weights to generate robust implicit features $\mathbf{v}^l$. We then integrate these multi-resolution features $\{\mathbf{v}^l\}_{l=1}^L$ through a top-down attention mechanism (Section~\ref{sec:adaptive_integration}), which sequentially refines and combines feature representations from coarse to fine resolutions. Finally, we decode the consolidated feature representations $\mathbf{v}^{\text{HR}}$ into RGB values using a multi-layer perceptron (MLP), generating the final HR video frames $\mathbf{I}^{\text{HR}}_{t}$.

\subsection{Spatial-Temporal-Texture Encoding}
\label{sec:texture_encoding}

We first extract multi-scale local patches from each LR frame guided by hierarchical resolution grids $\{\mathbf{r}^{l}\}_{l=1}^L$ . Specifically, given a target high-resolution coordinate $\mathbf{r}^{\text{HR}}$, we first resize the LR frames to the target resolution using bicubic interpolation, and query the local patches at various resolutions by:
\begin{equation}
\mathbf{P}_i^l = \mathbf{I}^{\text{LR}}(\mathbf{r}^{l}_i),
\end{equation}
where $\mathbf{P}_i^l$ denotes the local patch at resolution level $l$, as shown in Fig.~\ref{fig:pipeline}.
These patches are then encoded into compact texture feature representations:
\begin{equation}
\label{eqn:patch_enc}
\mathbf{T}^l_i = \mathcal{G}_{\text{T}}^l(\mathbf{P}^l_i),
\end{equation}
where $\{\mathcal{G}_{\text{T}}^l\}_{l=1}^L$ is a set of MLPs to map the flattened patches of different resolution to fixed-length texture codes $\mathbf{T}^l  = [\boldsymbol{\xi}^l_1,\boldsymbol{\xi}^l_2,...\boldsymbol{\xi}^l_F]$, resulting in hierarchical feature representations across multiple resolutions. In our implementation, we use a three-dimensional feature code, \ie $F=3$, to represent the local texture information.
At each resolution level $l$, we then concatenate the feature codes $\mathbf{T}^l$ to the query HR spatial-temporal coordinate $\mathbf{r}^{\text{HR}}_t = [\mathbf{x}, \mathbf{y}, \mathbf{t}]$ to obtain a spatial-temporal-texture (STT) coding representation:
\begin{equation}
\label{eqn:stt}
    \mathbf{F}^l_{\text{STT}}(\mathbf{r}^{\text{HR}}_t) = [\mathbf{x}, \mathbf{y}, \mathbf{t}, \boldsymbol{\xi}^l_{1},\boldsymbol{\xi}^l_2,...\boldsymbol{\xi}^l_F].
\end{equation}


\subsection{Implicit Feature Interpolation via Hashing}
\label{sec:implicit_hashing}

To generate robust implicit multi-resolution features from the spatial-temporal-texture (STT) codes $\mathbf{F}_{\text{STT}}^l$, we utilize implicit hashing to efficiently interpolate features stored within a spatial hash table. Each STT code $[\mathbf{x}, \mathbf{y}, \mathbf{t}, \boldsymbol{\xi}^l_{1},\boldsymbol{\xi}^l_2,...\boldsymbol{\xi}^l_F]$($F=3$), is represented as a 6-dimensional vector in which each dimension ranges between [-1,1]. Given the spatial resolution grid $\mathbf{r}^{l}$ at resolution level $l$, we partition the 6-dimensional feature space accordingly, identifying the vertices nearest to each STT code $\hat{\mathbf{F}}_{\text{STT}}^l$.
For example, along the first dimension $\mathbf{x}$, the neighboring vertices can be defined as:
\begin{equation}
\mathbf{x}_{\text{min}}^l = \lfloor \hat{\mathbf{x}}^l \rfloor, \quad \mathbf{x}_{\text{max}}^l = \lceil \hat{\mathbf{x}}^l \rceil,
\end{equation}
where $\hat{\mathbf{x}}^l$ is the normalized spatial coordinate of the STT code in the $\mathbf{x}$ dimension. Similarly, neighboring vertices are identified along dimensions $\mathbf{y}$, $\mathbf{t}$, and texture dimensions $\boldsymbol{\xi}_f^l$, for $f=1,2,3$. Consequently, we identify all $2^6=64$ neighboring vertices $\{\mathbf{V}_n|n=1,...64\}\in\mathbb{R}^6$ around the target STT code in the 6-dimensional latent space.
To enhance training and inference efficiency, we retrieve the corresponding feature vectors from the hash table:
\begin{equation}
\hat{\mathbf{v}}_{n}^l = \text{HashTable}(\mathbf{V}_n).
\end{equation}

Unlike methods such as Instant-NGP \cite{mueller2022instant}, which use simple interpolation methods, we propose an implicit interpolation method utilizing learned adaptive weights to combine neighboring hashed features. The adaptive interpolation weights are predicted using a dedicated network $\mathcal{G}_{\text{Hash}}^l$ based on the relative position of the input STT code within its 6-dimensional neighborhood:
\begin{equation}
\label{eqn:hashnet}
[\mathbf{w}_{1}^l, \dots, \mathbf{w}_{64}^l] = \mathcal{G}_{\text{Hash}}^l\left(\mathbf{F}_{\text{STT}}^l, \mathbf{F}_{\text{STT}}^l - \mathbf{V}_{\text{min}}^l, \mathbf{V}_{\text{max}}^l - \mathbf{F}_{\text{STT}}^l\right),
\end{equation}
where $\mathbf{V}_{\text{min}}^l$ and $\mathbf{V}_{\text{max}}^l$ represent the boundary vertices in the 6-dimensional latent space. Finally, the interpolated implicit feature vector $\mathbf{v}^l$ at resolution level $l$ is computed as a weighted combination:
\begin{equation}
\mathbf{v}^l = \sum_{n=1}^{64} \mathbf{w}_{n}^l \cdot \hat{\mathbf{v}}_{n}^l.
\end{equation}

\subsection{Adaptive Multi-Resolution Feature Integration}
\label{sec:adaptive_integration}
To effectively integrate features across multiple resolutions, we propose a top-down attention mechanism that adaptively refines feature representations from coarser (larger patch areas) to finer (smaller patch areas) resolution layers, as shown in Fig.~\ref{fig:pipeline}. Specifically, starting from the coarsest resolution level $L$, we compute attention weights at each subsequent finer resolution level. Formally, for each level $l$ ($1 \leq l < L$), the attention weights are computed using features from the immediately coarser resolution level ($l+1$):

\vspace{-10pt}
\begin{equation}
\label{eqn:attnet}
\mathbf{w}_{\text{att}}^{l} = \mathcal{G}_{\text{att}}^l\left(\mathbf{v}^{l+1}\right),
\end{equation}
where $\mathcal{G}_{\text{att}}^l$ is a dedicated two-layer MLP designed to generate adaptive weights based on features from the larger patch area at resolution level $(l+1)$.
We then explicitly multiply these computed attention weights with the corresponding finer-resolution features to obtain refined feature representations:
\begin{equation}
\mathbf{v}^{l} = \mathbf{w}_{\text{att}}^{l} \odot \mathbf{v}^{l},
\end{equation}
where $\odot$ denotes element-wise multiplication. Consequently, coarser-resolution features provide context-aware guidance for iteratively refining finer-resolution features.
After applying attention-based refinement across all resolution layers, we concatenate the adaptively integrated features from each resolution level to form the final multi-resolution feature vector:
\begin{equation}
\mathbf{v}^{\text{HR}} = [\mathbf{v}^1, \mathbf{v}^2, \dots, \mathbf{v}^L].
\end{equation}

Finally, we decode the concatenated multi-resolution feature vector into the final RGB color value using a two-layer MLP:
\begin{equation}
\label{eqn:colornet}
\hat{\mathbf{I}}^{\text{HR}} = \mathcal{G}_{\text{color}}(\mathbf{v}^{\text{HR}}),
\end{equation}
where $\mathcal{G}_{\text{color}}$ is an MLP with one hidden layer. A detailed network architecture of all the MLPs can be found in Section.~\ref{sec:details}.

\subsection{Training Details.}
\label{sec:pea}
Due to the pixel-based encoding and decoding nature of our method, directly using mean squared error (MSE) loss may lead to over-smoothed reconstructions, as it equally penalizes all pixel errors. This can cause the network to neglect subtle yet important details, especially in regions with low reconstruction errors. To mitigate this, we propose a novel Pixel-Error Amplified Loss (PEA-loss).
First, we calculate the standard per-pixel reconstruction error:
\begin{equation}
\mathcal{L}_{\text{pixel}} = \text{MSE}(\hat{\mathbf{I}}^{\text{HR}}, {\mathbf{I}}^{\text{HR}}),
\end{equation}
where $\hat{\mathbf{I}}^{\text{HR}}$ is the reconstructed HR frames, and ${\mathbf{I}}^{\text{HR}}$ is the ground-truth HR frames.
We then apply a reconstruction mask, \(M_{\text{recon}}\), initialized to ones, and subsequently updated during training. Pixels with errors smaller than a predefined threshold \(\tau\) are masked out, preventing the model from overly focusing on already well-reconstructed regions:
\begin{equation}
\mathbf{M}_{\text{recon}} = 
\begin{cases}
1, & \text{if } \mathcal{L}_{\text{pixel}} > \tau, \\
0, & \text{otherwise},
\end{cases}
\end{equation}
where $\tau$ is a predefined threshold. Importantly, rather than updating this mask iteratively, we keep the threshold fixed during training to ensure stable convergence.
The masked reconstruction loss is computed as:
\begin{equation}
\mathcal{L}_{\text{masked}} = \text{mean}(\mathcal{L}_{\text{pixel}} \odot \mathbf{M}_{\text{recon}}).
\end{equation}
To further refine subtle details in regions of lower error, we define an additional boosted loss component:
\begin{equation}
\mathcal{L}_{\text{boost}} = \mathcal{L}_{\text{pixel}} + \delta \cdot \mathbf{1}(\mathcal{L}_{\text{pixel}} < \epsilon),
\end{equation}
where $\mathbf{1}(\cdot)$ denotes the indicator function, adding a small constant $\delta$ only to pixels whose reconstruction errors are below the threshold $\epsilon$. This approach ensures that boosting specifically targets low-error regions to enhance detail preservation.
The final PEA-loss combines both masked reconstruction and boosted terms:
\begin{equation}
\label{eqn:pea}
\mathcal{L}_{\text{PEA}} = \mathcal{L}_{\text{masked}} + \alpha\cdot\mathcal{L}_{\text{boost}},
\end{equation}
where $\alpha$ controls the influence of the boosted loss.

\subsection{Inference}
\label{sec:inference}

At test time, given a degraded low-resolution (LR) video \(\{\hat{\mathbf{I}}^{\text{LR}}_{t}\}_{t=1}^T\), VR-INR first resizes $\hat{\mathbf{I}}^{\text{LR}}_{t}$ to the target resolution using bicubic interpolation, then restores to high-resolution (HR) as follows:

\begin{equation}
\label{eq:vr-inr-inference}
\begin{aligned}
\hat{\mathbf{I}}^{\mathrm{HR}}_t (\mathbf{r}^{\text{HR}}_t )
&= \mathcal{G}_{\mathrm{color}}\Bigl(\,
    \underbrace{\mathcal{G}_{\mathrm{att}}^1\!\bigl(\mathbf{v}^2\bigr)\odot\mathbf{v}^1}_{l=1}
    \;\big\|\;\dots\;\big\|\;
    \underbrace{\mathcal{G}_{\mathrm{att}}^L\!\bigl(\mathbf{v}^{L+1}\bigr)\odot\mathbf{v}^L}_{l=L}
\Bigr),
\\
\mathbf{v}^l
&= \sum_{n=1}^{64} w^l_n \;\mathrm{Hash}\!\Bigl(
       [\mathbf{r}^{\text{HR}}_t, \mathcal{G}^l_{\mathrm{T}}\bigl(\hat{\mathbf{I}}^{\text{LR}}(\mathbf{r}^l)\bigr)],\,
      \mathbf{F}^l_{\mathrm{STT}} - \mathbf{V}^l_{\min},\,
      \mathbf{V}^l_{\max} - \mathbf{F}^l_{\mathrm{STT}}
    \Bigr).
\end{aligned}
\end{equation}

VR-INR naturally supports \emph{any} spatial upscaling factor at inference and performs \emph{zero‑shot} denoising on noisy inputs without additional training, owing to the continuous implicit representation and learned hash‑encoding priors.


\section{Experiments}

\subsection{Implementation Details} 
\label{sec:details}
All the networks used are two-layer Multi-Layer Perceptrons (MLPs) with ReLU activation functions and hidden layer dimensions of 64 units. We used the Adam optimizer with an initial learning rate of $0.0001$. The learning rate was reduced by a factor of 0.5 every 100 epochs. All experiments were conducted on an NVIDIA A100 GPU. The network architecture details of VR-INR is as follows:

\noindent \textbf{Hierarchical Texture Encoding Network ($\mathcal{G}_{\text{T}}^l$, Eqn.~\ref{eqn:patch_enc}).}
Each local patch extracted from LR frames is encoded into spatial-temporal-texture (STT) codes using a two-layer MLP. The network has a hidden layer size of 64 units, followed by ReLU activation, and outputs a 3-dimensional texture code within the range $[-1,1]$.

\noindent \textbf{Implicit Hashing Network ($\mathcal{G}_{\text{Hash}}^l$, Eqn.~\ref{eqn:hashnet}).}
The implicit hashing module employs a two-layer MLP with 64 hidden units and ReLU activation. Given a 6-dimensional STT code, this network predicts 64 interpolation weights corresponding to the neighboring vertices in the hash table.

\noindent \textbf{Top-Down Attention Network ($\mathcal{G}_{\text{att}}^l$, Eqn.~\ref{eqn:attnet}).}
The attention mechanism employs a two-layer MLP with 64 hidden units, using ReLU activation. It computes adaptive attention weights at each resolution level based on feature representations from the immediately coarser resolution, which are applied to refine finer-resolution features iteratively.

\noindent \textbf{Color Decoding Network ($\mathcal{G}_{\text{color}}$, Eqn.~\ref{eqn:colornet}).}
The final RGB values for high-resolution reconstruction are predicted using a two-layer MLP with 64 hidden units and ReLU activation, mapping concatenated multi-resolution feature representations to RGB outputs.

\noindent \textbf{Pixel-Error Amplified Loss (PEA-Loss) Hyperparameters. (Sec.~\ref{sec:pea})} For the proposed Pixel-Error Amplified Loss, we set the reconstruction error threshold ($\tau$) to 0.01, the boosting error threshold ($\epsilon$) to 0.005, the boosting constant ($\delta$) to 0.001, and the boosting weight factor ($\alpha$) to 5. A detailed analysis and justification of these hyperparameters are provided in the ablation studies.

\subsection{Comparison Experiments}
\paragraph{Dataset}  
We adopt four widely used video datasets in experiments, \ie Vid4 \cite{Vid4}, REDS4 \cite{nah2019ntire}, GOPRO \cite{gopro}, and DAVIS \cite{pont20172017}. 
For super-resolution (SR) evaluation, LR images were generated by bicubic downsampling of HR images at these scaling factors to simulate varying degrees of image degradation. For DAVIS and GOPRO, we first resized the video frames to $256 \times 256$ pixels, which served as the HR ground truth. Our model was primarily trained on $\times 4$ scaling and evaluated on both in-distribution ($\times 4$) and out-of-distribution ($\times 2 \sim 32$) scales. 

\begin{figure}[ht]
    \centering
    \includegraphics[width=1\textwidth]{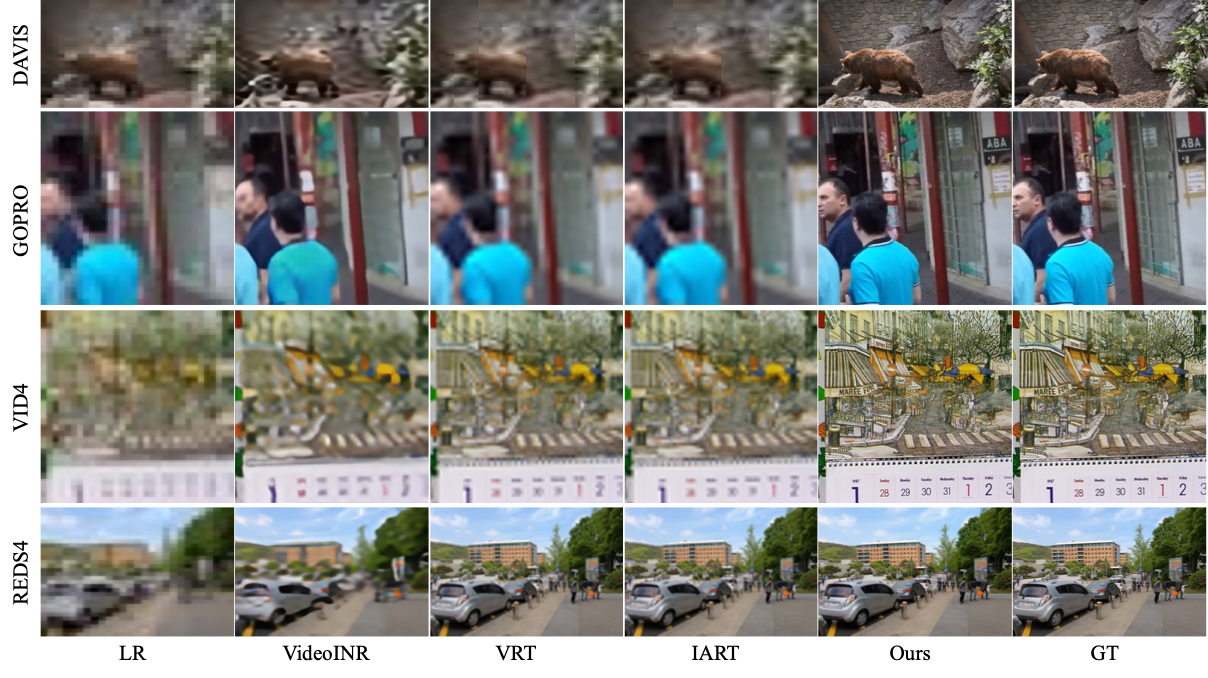}
    \vspace{-7pt}
    \caption{Visual comparison with state-of-the-art methods on the Vid4 \cite{Vid4}, REDS4 \cite{nah2019ntire}, GOPRO \cite{gopro}, and DAVIS \cite{pont20172017} datasets for video super-resolution at $\times 8$ (unseen) scaling factors.}
    \label{fig:qual_figure}
    \vspace{-10pt}
\end{figure}

\paragraph{Compared with SOTAs}
We compare VR‑INR against several leading video restoration and super‑resolution techniques, including VRT \cite{liang2024vrt}, VideoINR \cite{chen2022videoinr}, IART \cite{xu2024enhancing}, and SAVSR\cite{li2024savsr}. We evaluate both \textbf{arbitrary upscaling} ($\times2 \sim 32$) and \textbf{zero‑shot denoising} (with additive Gaussian noise $\sigma=30, 50$). All baselines are pre‑trained exclusively at the $\times4$ setting and cannot natively handle other scales or noise without retraining. For quantitative comparison, we evaluate the video quality by PSNR and SSIM, as shown in Table \ref{tab:videoresult}. We also show the visual comparison in Fig.~\ref{fig:qual_figure}.

\begin{table*}[t!]

\caption{Quantitative comparison on Vid4 \cite{Vid4}, REDS4 \cite{nah2019ntire}, GOPRO \cite{gopro} and DAVIS \cite{pont20172017} dataset with the state-of-the-art for $\times 2$, $\times 4$ and $\times 8$ video SR scales. The best result is highlighted in \textbf{bold} and \blue{underline} texts respectively.} 
\centering
\setlength{\tabcolsep}{9pt}
  \resizebox{1\linewidth}{!}{ 
\begin{tabular}{l c c c c c c c c c}
\toprule
\multirow{2}{*}{Scale} & \multirow{2}{*}{Methods} & \multicolumn{2}{c}{VID4} & \multicolumn{2}{c}{REDS4} & \multicolumn{2}{c}{GOPRO}& \multicolumn{2}{c}{DAVIS}\\
& & PSNR & SSIM & PSNR & SSIM & PSNR & SSIM & PSNR & SSIM\\
\midrule
\multirow{4}{*}{$\times$2} 
& VideoINR \cite{chen2022videoinr} & {28.08} & {0.851} & {25.07} & {0.777}  & {26.00} & {0.822} &  {27.10} & {0.798} \\
& VRT \cite{liang2024vrt} & {-} & {-} & {-} & {-} & {-} & {-} &  {-} & {-} \\
& IART \cite{xu2024enhancing} & {-} & {-} & {-} & {-}  & {-} & {-} & {-} & {-}\\
& SAVSR \cite{li2024savsr} & \blue{30.95} & \blue{0.937} & \blue{33.34} & \blue{0.949}  & \textbf{37.40} & \textbf{0.976} & \blue{35.82} & \blue{0.960}\\
& VR-INR (ours) & \textbf{43.68} & \textbf{0.990} & \textbf{35.03} & \textbf{0.953} & \blue{34.55} & \blue{0.948} &  \textbf{40.16} & \textbf{0.985} \\
\midrule
\multirow{4}{*}{$\times$4} 
& VideoINR \cite{chen2022videoinr} & {24.21} & {0.656} & {26.50} & {0.770} & {28.96} & {0.842} &  {24.69} & {0.698} \\
& VRT \cite{liang2024vrt} & {27.93} & {0.843} & {32.19} & {0.901}  & {28.80} & {0.854} &  {26.37} & {0.703} \\
& IART \cite{xu2024enhancing} & \blue{28.26} & \blue{0.852} & \blue{32.90} & \blue{0.914}  & {32.22} & {0.924} & {26.35} & {0.703} \\
& SAVSR \cite{li2024savsr} & {24.50} & {0.718} & {27.14} & {0.811}  & \blue{30.16} & \blue{0.881} & \blue{30.10} & \blue{0.861}\\
& VR-INR (ours) & \textbf{44.21} & \textbf{0.996} & \textbf{36.79} & \textbf{0.977} & \textbf{36.50} & \textbf{0.975} &  \textbf{42.00} & \textbf{0.984} \\
\midrule
\multirow{4}{*}{$\times$8}
& VideoINR \cite{chen2022videoinr} & \blue{20.67} & {0.479} & {22.02} & \blue{0.618} & {23.65} & \blue{0.707}  &  {21.69} & {0.631} \\
& VRT \cite{liang2024vrt} & {21.31} & {0.469} & {24.01} & {0.596} & {23.03} & {0.575} &  {22.83} & {0.563} \\
& IART \cite{xu2024enhancing} & {21.45} & \blue{0.482} & \blue{24.12} & {0.598}  & {22.97} & {0.574} & {22.82} & {0.562}\\
& SAVSR \cite{li2024savsr} & {21.36} & {0.461} & {23.30} & {0.597}  & \blue{25.38} & {0.693} & \blue{25.44} & \blue{0.690}\\
& VR-INR (ours) & \textbf{41.42} & \textbf{0.985} & \textbf{33.78} & \textbf{0.930} &  \textbf{33.30} & \textbf{0.917}  &  \textbf{40.46} & \textbf{0.970} \\
\bottomrule
\end{tabular}
\vspace{-10pt}
}
\label{tab:videoresult}
\end{table*}

\begin{figure}[htb]
    \centering
    \begin{minipage}{0.53\linewidth}
        \includegraphics[width=.9\linewidth]{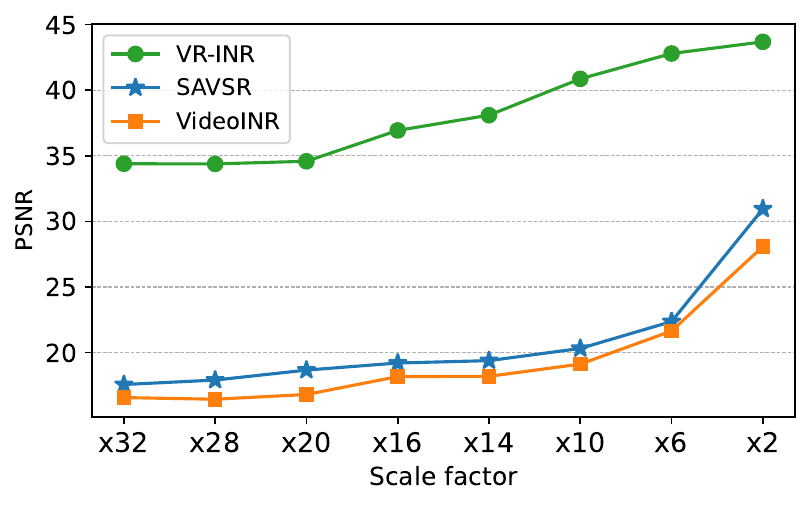}
    \vspace{-5pt}
        \captionof{figure}{The video super resolution effectiveness of our model for various arbitrary scales on Vid4 \cite{Vid4}.}
        \label{fig:scales}
    \end{minipage}
    \begin{minipage}{0.41\linewidth}
        \centering
        \captionof{table}{PSNR results on zero-shot denoising at noise levels $\sigma = 30$ and $\sigma = 50$, and super-resolution scale factors of $\times 4$ and $\times 8$ on the DAVIS dataset.}
        \small
        \setlength{\tabcolsep}{5pt}
        \begin{tabular}{lcccc}
            \toprule
            \multirow{2}{*}{Method} & \multicolumn{2}{c}{$\sigma=30$} & \multicolumn{2}{c}{$\sigma=50$} \\
            \cmidrule(lr){2-3} \cmidrule(lr){4-5}
            & $\times4$ & $\times8$ & $\times4$ & $\times8$ \\
            \midrule
            VideoINR  & {18.11} & {16.87} & {14.86} & {13.97} \\
            SAVSR  & {19.88} & {19.48} & {16.64} & {16.62} \\
            VRT  & {18.70} & {17.94} & {14.92} & {14.61} \\
            VR-INR  & \textbf{31.50} & \textbf{31.22} & \textbf{30.68} & \textbf{30.62} \\
            \bottomrule
        \end{tabular}
        \label{tab:zero_shot_denoising}
    \end{minipage}
    \vspace{-10pt}
\end{figure}
    \vspace{-10pt}




\begin{figure}[ht]
    \centering
    \includegraphics[width=1\textwidth]{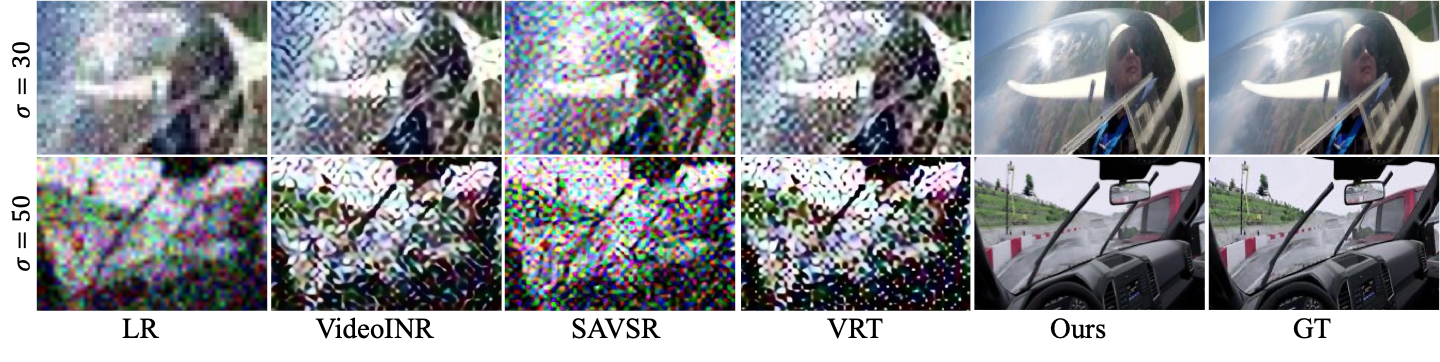}
    \vspace{-10pt}
\caption{Visual comparison with state-of-the-art (SOTA) video super-resolution (VSR) methods on zero-shot denoising under Gaussian noise at $\sigma=30$ and $\sigma=50$, evaluated at a $\times 4$ input resolution on the DAVIS~\cite{pont20172017} dataset.}

    \label{fig:denoise}
    \vspace{-10pt}
\end{figure}

\paragraph{Evaluation on Video Super Resolution.}
Table~\ref{tab:videoresult} presents quantitative comparisons demonstrating our method’s effectiveness across multiple benchmarks and scaling factors. To ensure a fair and consistent evaluation, we first measured their performance at this trained scale and subsequently tested their generalization at untrained scales (\(\times 2\) and \(\times 8\)). In particular, methods such as VRT and IART could not be evaluated on the scale \(\times 2\) due to limitations in their original design. In Fig.~\ref{fig:scales}, we compare the PSNR curves of our method with VideoINR and SAVSR on arbitrary SR scales.

\paragraph{Evaluation on Zero-shot Denoising}
We assess VR‑INR’s ability to remove noise without any noise‑specific training by comparing against VideoINR~\cite{chen2022videoinr}, SAVSR~\cite{li2024savsr}, IART~\cite{xu2024enhancing}, and VRT~\cite{liang2024vrt}. All baselines are pre‑trained solely for super‑resolution using clean LR–HR pairs and have never been exposed to noisy inputs. Despite this, VR‑INR consistently outperforms these methods in both PSNR and SSIM on noisy test sequences. Quantitative results are presented in Table~\ref{tab:zero_shot_denoising}, and visual examples are shown in Fig.~\ref{fig:denoise}.

\begin{figure}[htb]
    \centering
    \begin{minipage}{0.54\linewidth}
        \centering
        \includegraphics[width=.95\linewidth]{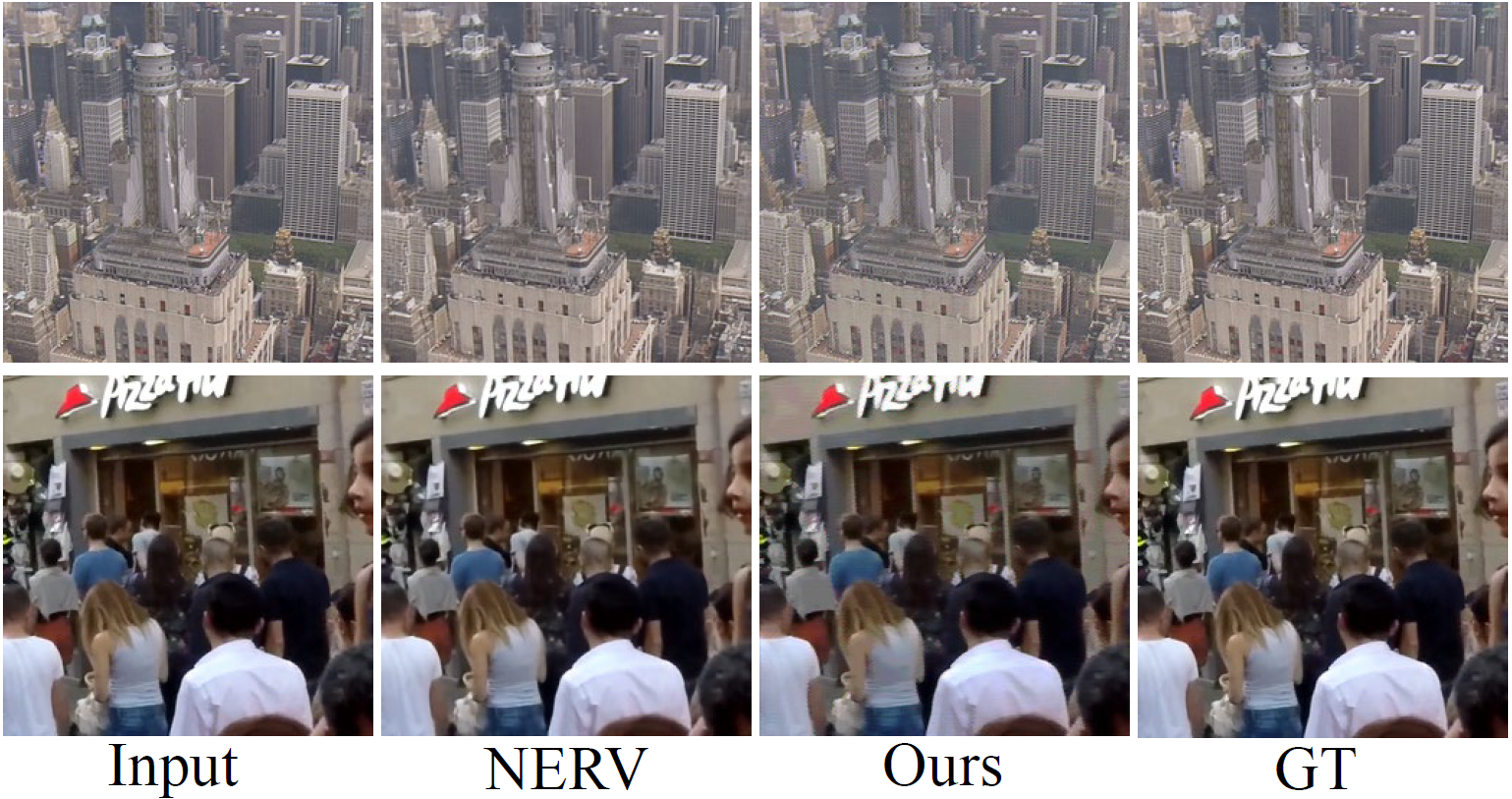}
    \vspace{-5pt}
        \captionof{figure}{Results of NERV and Ours. From top to bottom: Vid4 and GOPRO datasets.}
        \label{fig:nerv}
    \end{minipage}
    \hfill
    \begin{minipage}{0.42\linewidth}
        \centering
        \captionof{table}{Quantitative comparison on video datasets including Vid4 and GOPRO. The best result in PSNR and SSIM is highlighted in bold.}
        \small
        \setlength{\tabcolsep}{4pt}
        \begin{tabular}{lcccc}
            \toprule
            \multirow{2}{*}{Method} & \multicolumn{2}{c}{Vid4} & \multicolumn{2}{c}{GOPRO} \\
            \cmidrule(lr){2-3} \cmidrule(lr){4-5}
            & PSNR & SSIM & PSNR & SSIM \\
            \midrule
            NERV  & 35.446 & 0.976 & 32.028 & 0.970 \\
            Ours  & \textbf{43.68} & \textbf{0.990} & \textbf{34.55} & \textbf{0.948} \\
            \bottomrule
        \end{tabular}
        \label{tab:quantitative_on_nerv}
    \end{minipage}
    \vspace{-10pt}
\end{figure}

\paragraph{Video Reconstruction}
We compare VR‑INR with NeRV~\cite{chen2021nerv}, a state‑of‑the‑art implicit video reconstruction model, on the GOPRO and VID4 datasets. For pure reconstruction and zero‑shot denoising (in Appendix), VR‑INR consistently achieves higher PSNR and SSIM than NeRV (Table~\ref{tab:quantitative_on_nerv}) and yields visibly sharper, more detailed frames (Fig.~\ref{fig:nerv}). These results demonstrate VR‑INR’s superior versatility in handling both faithful video reconstruction and denoising without any noise‑specific training.  

\subsection{Ablation Studies}
We conduct extensive ablation studies to investigate the impact of various architectural choices and hyperparameters on our model's performance. We carried out these studies on the DAVIS dataset with a scale factor of $\times 4$. We evaluate the model using PSNR and SSIM metrics. All experiments maintain consistent settings except for the specifically varied component. These findings provide valuable guidance for configuring the model architecture to achieve the desired balance between performance and computational efficiency.


\noindent
\textbf{Feature Codes Length (Eqn.~\ref{eqn:stt}).}
Table \ref{tab:featuresperlevel} presents the impact of varying the feature code length per level in the hash table. The results indicate that using a feature code length of $6$ achieves the highest PSNR ($46.93$dB) while maintaining a high SSIM. 

\vspace{-10pt}
\begin{table}[h!]
  \centering
  \begin{minipage}{0.48\linewidth}
    \caption{Ablation study on the number of feature codes length $F$ in the hash table. The best PSNR and SSIM results are in bold.}
    \centering
    \small
    \setlength{\tabcolsep}{6pt}
    \begin{tabular}{ccccc}
      \toprule
      & 4 & 5 & 6 & 7 \\
      \midrule
      PSNR & 44.34 & 40.85 & \textbf{46.93} & 35.21 \\
      SSIM & \textbf{0.995} & 0.989 & 0.993 & 0.988 \\
      \bottomrule
    \end{tabular}
    \label{tab:featuresperlevel}
  \end{minipage}%
  \hfill
  \begin{minipage}{0.48\linewidth}
    \caption{Study on impact of attention mechanism and total $\mathcal{L}_{\text{PEA}}$ loss on model performance.}
    \centering
    \small
    \begin{tabular}{c c c | c c}
      \toprule
      Attention & $\mathcal{L}_{\text{pixel}}$ & $\mathcal{L}_{\text{PEA}}$ & PSNR & SSIM \\
      \midrule
       & \checkmark & \checkmark & 33.51 & 0.937 \\
      \checkmark & \checkmark &  & 38.59 & 0.976 \\
      \checkmark & \checkmark & \checkmark & \textbf{46.93} & \textbf{0.993} \\
      \bottomrule
    \end{tabular}
    \label{tab:attention_loss}
  \end{minipage}
    \vspace{-10pt}
\end{table}

\begin{figure}[htb]
    \centering
    \includegraphics[width=0.65\textwidth]{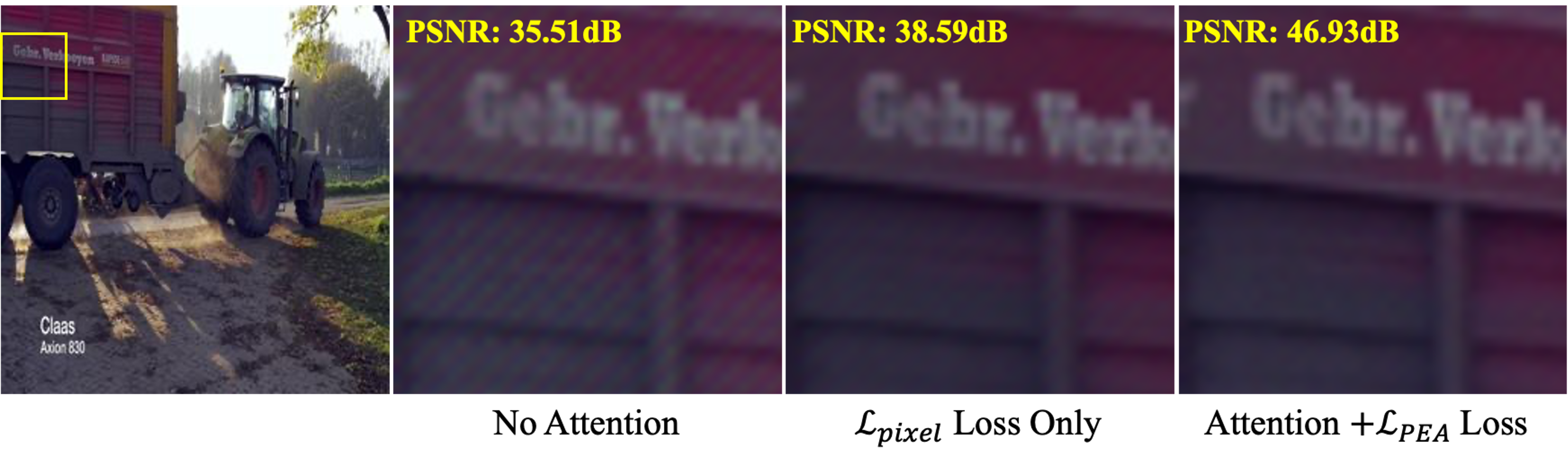}
    \vspace{-10pt}
   \caption{Visual comparison of the impact of our top-down attention mechanism and the $\mathcal{L}_{\text{PEA}}$ loss on super-resolution quality.}
    \label{fig:abl}
    \vspace{-10pt}
\end{figure}

\paragraph{Top-Down Attention Mechanism (Sec.~\ref{sec:adaptive_integration}):} To assess the contribution of the top-down attention mechanism, we performed an ablation study by removing the attention component from the feature concatenation process. Without the attention mechanism, features from different resolution layers were directly concatenated without any prioritization. The results, shown in Table \ref{tab:attention_loss}, indicate a significant drop in both PSNR and SSIM. This is due to the model's reduced ability to effectively integrate information from multiple resolutions, leading to less refined feature representations and poorer texture consistency. As shown in Fig.~\ref{fig:abl}, the result without the attention mechanism exhibits grid-like artifacts and reduced visual clarity.



\paragraph{Pixel-Error Amplified Loss (PEA-Loss, Eqn.~\ref{eqn:pea}):} We evaluated the effectiveness of our proposed Pixel-Error Amplified Loss ($\mathcal{L}_{\text{PEA}}$) by conducting an ablation study in which the model was trained using only the standard per-pixel Mean Squared Error (MSE) loss, denoted as $\mathcal{L}_{\text{pixel}}$. Table \ref{tab:attention_loss} shows that the model trained with $\mathcal{L}_{\text{PEA}}$ achieved significantly better results in terms of PSNR and SSIM. The PEA-loss amplifies subtle reconstruction errors, allowing the model to focus on refining regions that would otherwise be neglected by the standard MSE loss, ultimately boosting performance. 
Furthermore, we investigate the effects of the hyperparameters in our proposed Pixel-Error Amplified Loss (PEA-loss): the reconstruction masking threshold (\(\tau\)), the error boosting threshold (\(\epsilon\)), the boosting constant (\(\delta\)), and the weight factor (\(\alpha\)).  Our experiments on the DAVIS dataset demonstrate that the model's performance remains relatively stable when \(\tau\), \(\epsilon\), and \(\delta\) are set within small ranges. Specifically, we observed minor performance variations when adjusting these three parameters, indicating that as long as they remain sufficiently small, their precise values do not substantially impact reconstruction quality. However, excessively increasing these thresholds can reduce effectiveness by either neglecting important pixels or unnecessarily amplifying trivial errors, which was confirmed by decreased performance when significantly larger values were tested.
The hyperparameter \(\alpha\), controlling the relative weight of the boosted loss term, has the most significant impact on the model's performance. Increasing \(\alpha\) effectively strengthens the emphasis on pixels with very low reconstruction errors, promoting finer detail reconstruction. Based on extensive experimentation, we selected \(\alpha = 5\), as this value provided an optimal balance between enhancing subtle details and maintaining stable training convergence.  

\section{Conclusion}
We have presented VR‑INR, a unified implicit neural representation framework for video restoration that simultaneously addresses super‑resolution and zero‑shot denoising. VR‑INR combines a hierarchical spatial–temporal–texture encoder, multi‑resolution hash encoding, and a top‑down attention mechanism to map degraded low‑resolution frames to high‑fidelity outputs at arbitrary scales ($\times$2–32) without retraining. By fine‑tuning per video using only clean LR–HR pairs, VR‑INR adapts to each sequence’s unique content and noise characteristics, delivering superior PSNR and SSIM on Vid4, REDS4, GOPRO, and DAVIS—even under unseen noise levels. Unlike traditional flow‑based or task‑specific networks, our approach is flow‑free, scale‑agnostic, and computationally efficient, simplifying the restoration pipeline. Future work will explore extending VR‑INR to temporal interpolation and further reducing inference time for real‑time deployment.

\small
\bibliographystyle{plain}
\bibliography{Neurips_main}

\newpage
\appendix
\appendix
\section{Visual Comparison}
\begin{figure}[htb]
    \centering
    \includegraphics[width=1\linewidth]{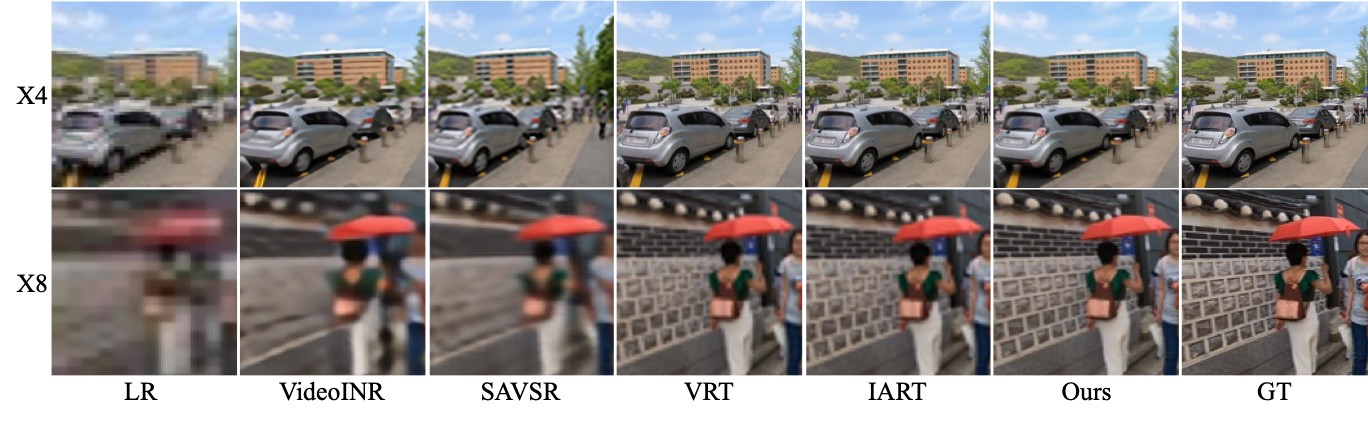}
        \captionof{figure}{Visual comparison of our model against state-of-the-art methods on the REDS4 dataset for scale factors of x4 and x8.}
        \label{fig:suppl_reds}
\end{figure}

\begin{figure}[htb]
    \centering
    \includegraphics[width=1\linewidth]{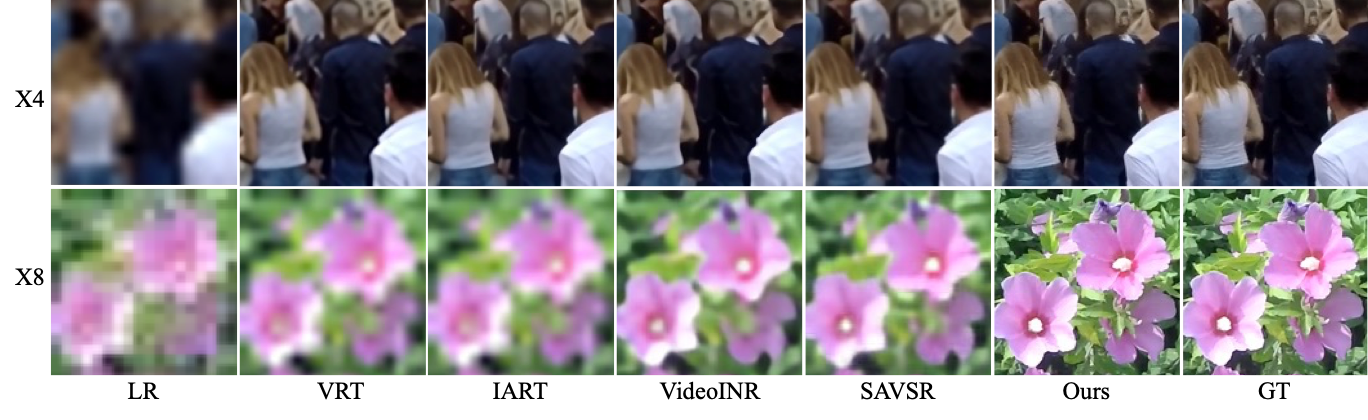}
        \captionof{figure}{Visual comparison of our model against state-of-the-art methods on the GOPRO dataset for scale factors of x4 and x8.}
        \label{fig:suppl_gopro}
\end{figure}

\begin{figure}[htb]
    \centering
    \includegraphics[width=.9\linewidth]{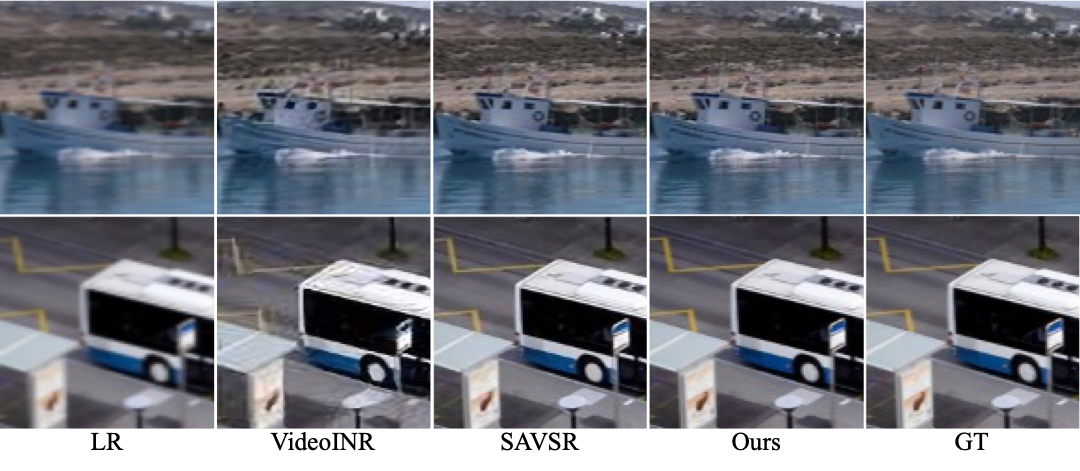}
        \captionof{figure}{Visual comparison of for x2 scale factor on the DAVIS dataset.}
        \label{fig:suppl_davis}
\end{figure}

We present additional qualitative comparisons with state-of-the-art (SOTA) video super-resolution methods, VRT \cite{liang2024vrt}, VideoINR \cite{chen2022videoinr}, IART \cite{xu2024enhancing}, and SAVSR \cite{li2024savsr} on the Vid4 \cite{Vid4}, REDS4 \cite{nah2019ntire}, GOPRO \cite{gopro} and DAVIS \cite{pont20172017} datasets. These comparisons span video super-resolution tasks at scale factors of ×2, ×4, and ×8, producing outputs at a resolution of 256×256.
As illustrated in Figure~\ref{fig:suppl_reds}, our model demonstrates superior preservation of fine textures and structural details on the REDS4 dataset. While other methods struggle to maintain edge clarity, resulting in blurred or smoothed patterns such as those in bricks and umbrellas, our model reconstructs sharp boundaries and detailed textures that closely resemble the ground truth.
On the GOPRO dataset in Fig. \ref{fig:suppl_gopro}, our model maintains visual fidelity even at the challenging ×8 scale. Other methods suffer from noticeable blurring, particularly in flower textures, whereas our model retains vibrant color and detail.
Fig. \ref{fig:suppl_davis} shows results on the DAVIS dataset at a ×2 scale.
These results highlight the effectiveness of our approach in reconstructing high-quality frames across varying datasets and scaling conditions, outperforming existing methods in terms of texture fidelity and edge sharpness. Video examples are available in the supplementary package.

\subsection{Arbitrary Scales}
We further evaluate the robustness of VR-INR under different arbitrary scales.  Fig. \ref{fig:scales} presents qualitative comparisons at scaling factors ranging from $\times4$ up to $\times32$ on the VID4 dataset. These experiments show the ability of each method to synthesize super-resolved frames from severely downsampled inputs.
As the scale increases, both VideoINR and SAVSR struggle to maintain spatial coherence, resulting in blurry and distorted outputs with significant detail loss. In contrast, VR-INR continues to generate sharper reconstructions with well-preserved textures and structure, even at $\times28$ and $\times32$. 
These results highlight the generalization ability of VR-INR, making it well-suited for applications that demand reliable frame synthesis under extremely low-resolution conditions.
\begin{figure}[htb]
    \centering
    \includegraphics[width=1\linewidth]{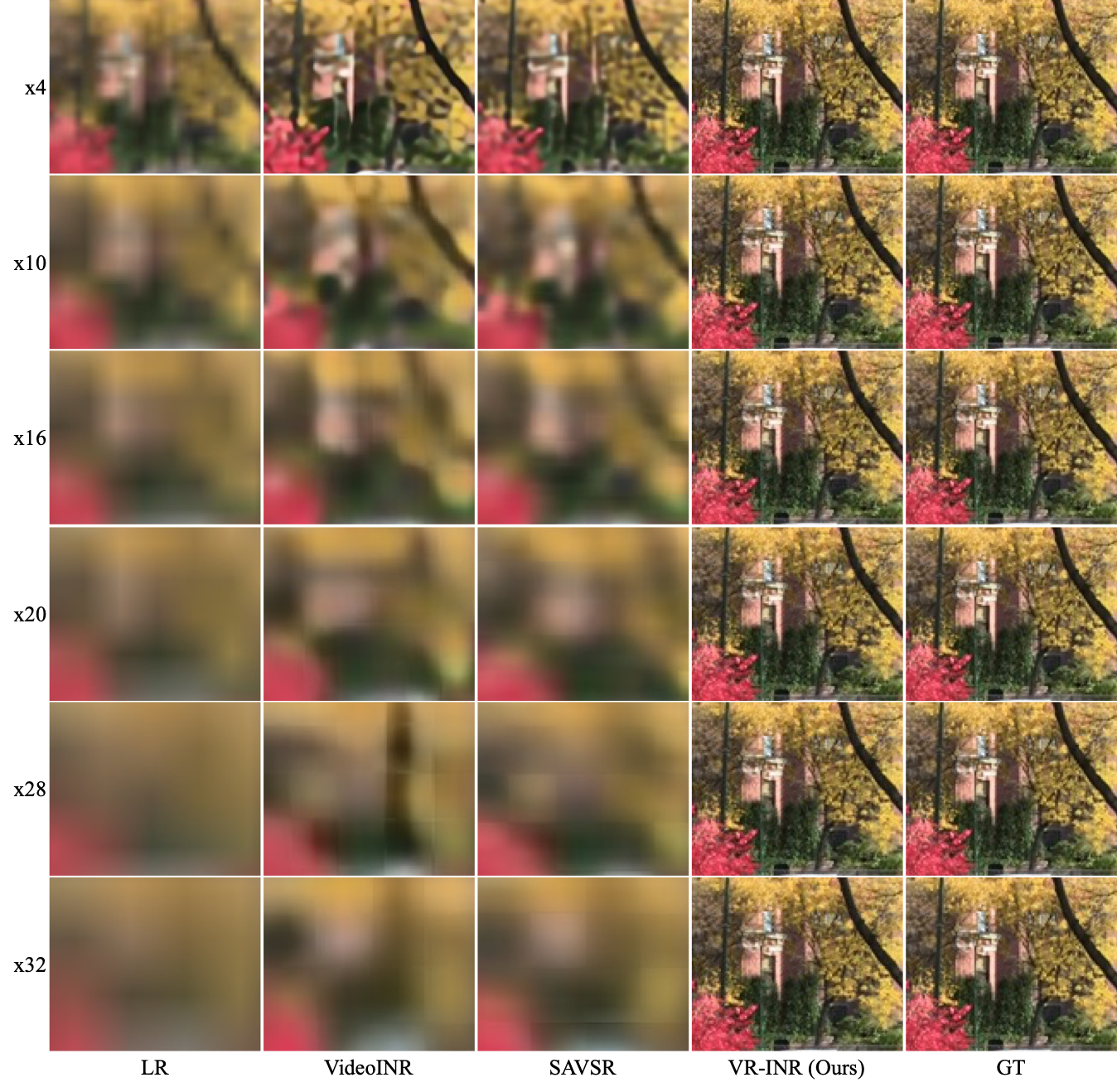}
        \captionof{figure}{Visual comparison of our model against state-of-the-art methods across various arbitrary scales on the Vid4 \cite{Vid4} dataset.}
        \label{fig:scales}
\end{figure}

\section{Zero-Shot Denoising}
To provide VR-INR’s generalization ability in performing zero-shot denoising with different noise conditions. While our model is originally designed for video super-resolution, we evaluate its effectiveness in handling noisy inputs without any retraining or noise-specific supervision. Specifically, we conduct experiments on the VID4 and DAVIS datasets corrupted with Gaussian noise (standard deviations of 30 and 50) and Poisson noise (intensity levels of 30 and 50). None of the models, including VR-INR and all baselines (VideoINR, VRT, IART, SAVSR), were trained with noisy inputs; they were optimized solely for super-resolution using clean low- and high-resolution frame pairs, without any exposure to noise during training.

\begin{figure}[htb]
    \centering
    \includegraphics[width=1\linewidth]{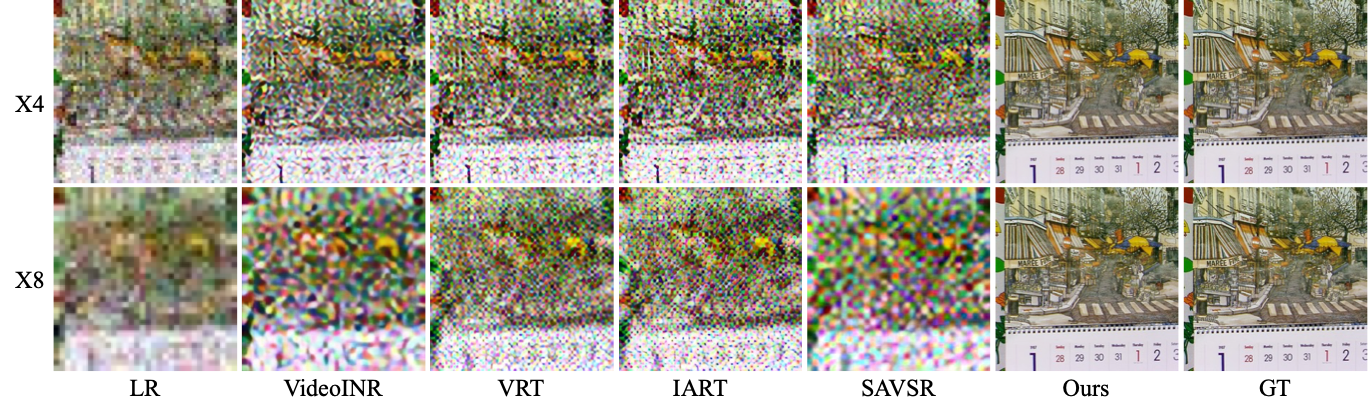}
        \captionof{figure}{Visual comparison to show the effectiveness of our model for performing zero-shot denoising for Gaussian 30}
        \label{fig:suppl_Gaussian30}
\end{figure}

\begin{figure}[htb]
    \centering
    \includegraphics[width=1\linewidth]{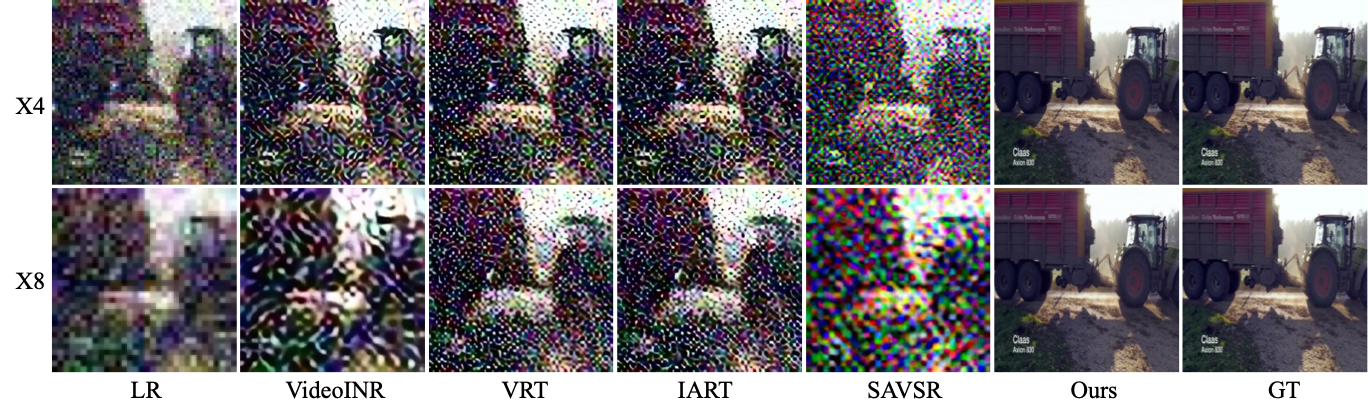}
        \captionof{figure}{Visual comparison to show the effectiveness of our model for performing zero-shot denoising for Gaussian 50}
        \label{fig:suppl_Gaussian50}
\end{figure}

Fig \ref{fig:suppl_Gaussian30} and Fig \ref{fig:suppl_Gaussian50} present qualitative comparisons under Gaussian noise with $\sigma = 30$ and $\sigma = 50$, respectively, across scale factors of $\times$4 and $\times$8. While baseline methods fail to remove noise and often produce severely distorted outputs, VR-INR demonstrates strong denoising ability, effectively recovering sharp textures and structures despite not being trained for this task.

\section{Reconstruction Ability Compared with NERV}
\begin{figure*}
    \centering
    \includegraphics[width=.9\textwidth]{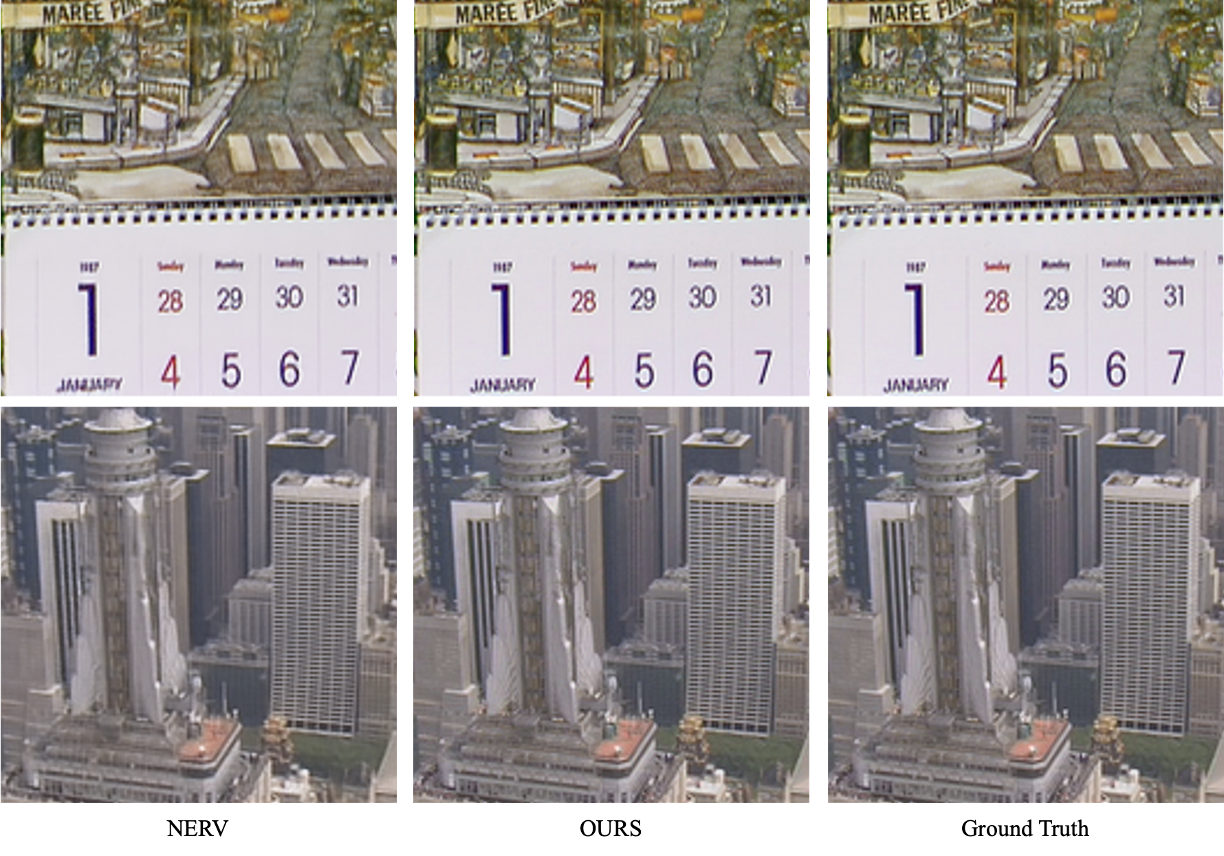}
    \caption{Visual comparison of video reconstruction on VID4 video dataset.}
    \label{div4_recon}
\end{figure*}

\begin{figure*}
    \centering
    \includegraphics[width=.9\textwidth]{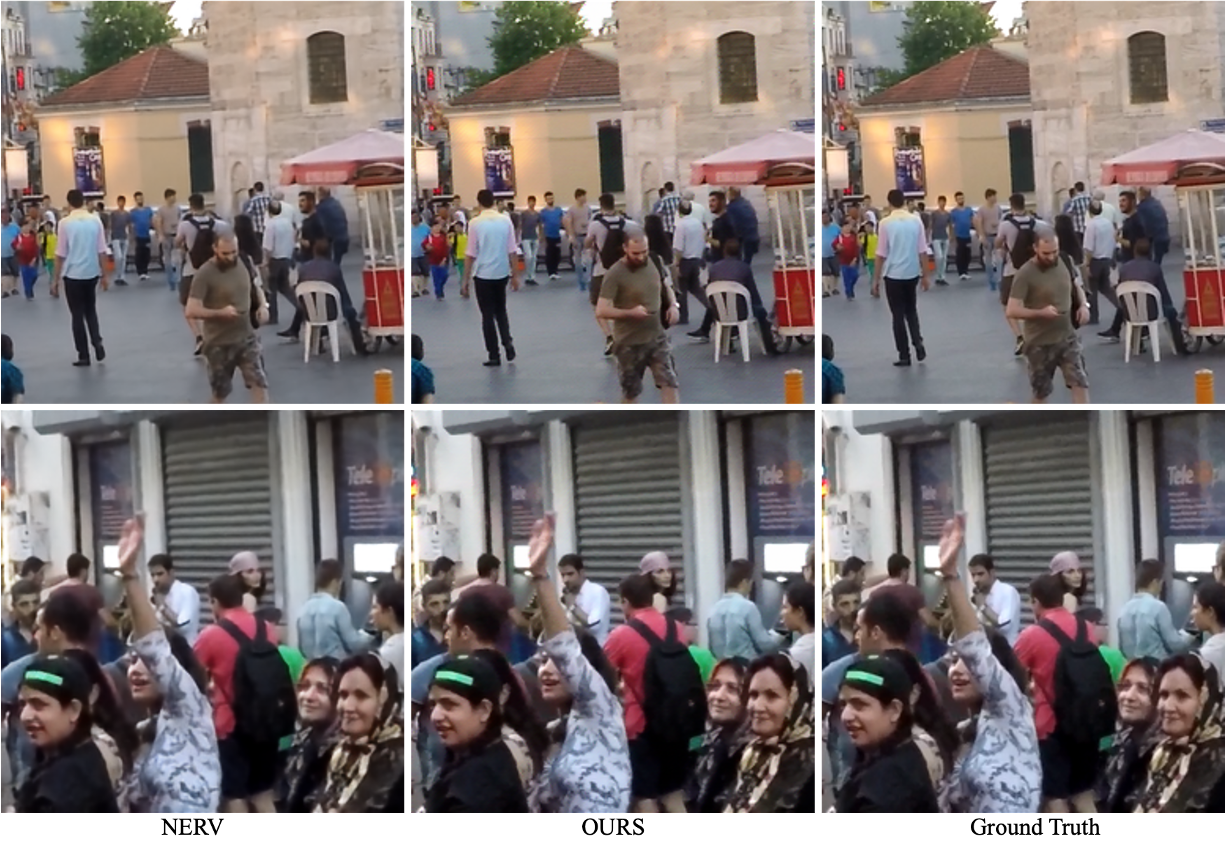}
    \caption{Visual comparison of video reconstruction on GOPRO video dataset.}
    \label{gopro_recon}
\end{figure*}

\begin{figure}[htb]
    \centering
    \includegraphics[width=1\linewidth]{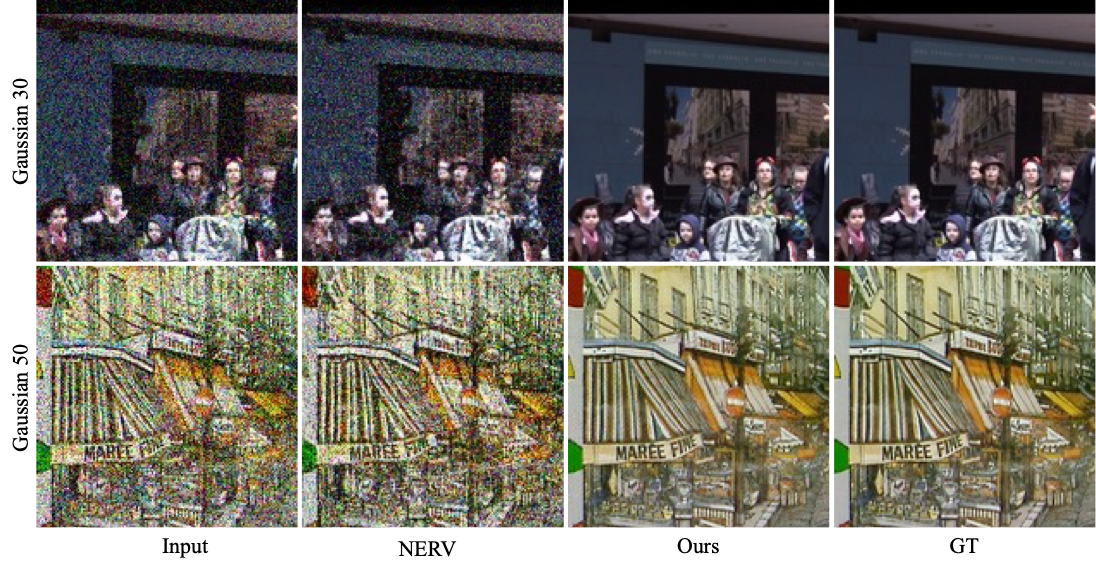}
        \captionof{figure}{Visual comparison of zero-shot denoising results using our video reconstruction framework under varying levels of Gaussian noise.}
        \label{fig:suppl_recon_gauss}
\end{figure}

\begin{figure}[ht]
    \centering
    \includegraphics[width=1\linewidth]{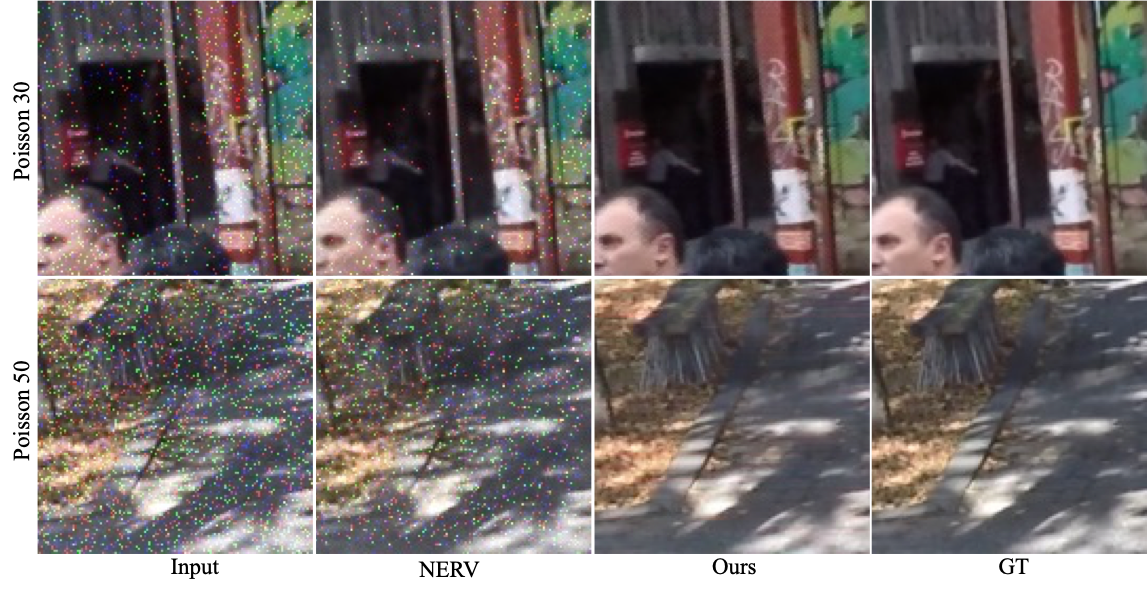}
        \captionof{figure}{Visual comparison of zero-shot denoising results using our video reconstruction framework under varying levels of Poisson noise.}
        \label{fig:suppl_recon_poiss}
\end{figure}

In addition to image and video super-resolution, our model demonstrates strong capabilities in video reconstruction tasks. To assess its reconstruction performance, we compared our method with NeRV, a state-of-the-art approach specifically designed for neural video representations, using the GOPRO and VID4 datasets. PSNR and SSIM metrics were used to quantify reconstruction quality.
As shown in Fig. \ref{div4_recon} and Fig. \ref{gopro_recon}, our model produces visually sharper and more detailed reconstructions that align closely with the ground truth. This demonstrates the versatility of our model in addressing a broader range of video-related tasks beyond its original super-resolution design.
Also, VR-INR is capable of performing zero-shot denoising in the context of video reconstruction. In this setting, the model is provided with noisy input sequences at their original resolution and tasked with reconstructing clean frames without any noise-specific training. We conducted experiments using Gaussian noise with standard deviations of $10, 30,$ and $50$, and Poisson noise at levels of $10, 30,$ and $50$. As shown in Fig. \ref{fig:suppl_recon_gauss} and Fig. \ref{fig:suppl_recon_poiss}, our model consistently suppresses noise while faithfully reconstructing the underlying video content, further underscoring its robustness in real-world degradation scenarios. The results presented in Table~\ref{tab:recon_noise} illustrate that our model outperforms NERV across different noise types and intensities.

\begin{table*}[h]
\centering
\caption{Reconstruction performance (PSNR/SSIM) of our method and NeRV on VID4 and GOPRO under different noise types and levels.}
\begin{tabular}{clcc|cc||cc|cc}
\toprule
\multirow{2}{*}{Noise Type} 
& \multirow{2}{*}{Level} 
& \multicolumn{2}{c|}{NeRV VID4} 
& \multicolumn{2}{c||}{NeRV GOPRO} 
& \multicolumn{2}{c|}{Ours VID4} 
& \multicolumn{2}{c}{Ours GOPRO} \\
& & PSNR & SSIM & PSNR & SSIM & PSNR & SSIM & PSNR & SSIM \\
\midrule
\multirow{3}{*}{Gaussian} 
& $\sigma=10$ & 29.37 & 0.879 & 29.43 & 0.802 & 41.91 & 0.973 & 33.36 & 0.9231 \\
& $\sigma=30$ & 20.05 & 0.559 & 20.11 & 0.425 & 39.33 & 0.926 & 32.83 & 0.9122 \\
& $\sigma=50$ & 15.55 & 0.358 & 15.78 & 0.251 & 37.13 & 0.913 & 32.10 & 0.8957 \\
\midrule
\multirow{3}{*}{Poisson}  
& $\lambda=10$ & 27.63 & 0.885 & 27.50 & 0.849 & 42.88 & 0.970 & 33.47 & 0.9260 \\
& $\lambda=30$ & 21.66 & 0.700 & 22.20 & 0.642 & 42.41 & 0.966 & 33.20 & 0.9192 \\
& $\lambda=50$ & 19.21 & 0.592 & 19.86 & 0.522 & 41.54 & 0.951 & 33.02 & 0.9129 \\
\bottomrule
\end{tabular}
\label{tab:recon_noise}
\end{table*}

\end{document}